\documentclass[10pt,twocolumn,letterpaper]{article}

\usepackage{cvpr}      

\usepackage[dvipsnames,table]{xcolor}
\usepackage{bbm}
\usepackage{diagbox}    
\usepackage{multirow}
\usepackage{booktabs}

\definecolor{mygray}{gray}{0.9}

\makeatletter
\newcommand*{\rom}[1]{\expandafter\@slowromancap\romannumeral #1@}
\makeatother

\definecolor{cvprblue}{rgb}{0.21,0.49,0.74}
\usepackage[pagebackref,breaklinks,colorlinks,citecolor=cvprblue]{hyperref}

\def\paperID{5286} 
\def\confName{CVPR}
\def\confYear{2024}

\title{Few-shot Learner Parameterization by Diffusion Time-steps}

\author{%
 \textbf{Zhongqi Yue}\textsuperscript{1,3}, \quad \textbf{Pan Zhou}\textsuperscript{2,3}, \quad \textbf{Richang Hong}\textsuperscript{4}, \quad \textbf{Hanwang Zhang}\textsuperscript{5,1}, \quad \textbf{Qianru Sun}\textsuperscript{2}\\
\small \textsuperscript{1}Nanyang Technological University,\quad \textsuperscript{2}Singapore Management University, \quad\small\textsuperscript{3}Sea AI Lab, \\\small\textsuperscript{4}Hefei University of Technology, \quad\small\textsuperscript{5} Skywork AI\\
\tt\small zhongqi.yue@ntu.edu.sg,\quad panzhou@smu.edu.sg,\quad hongrc.hfut@gmail.com,\\
\tt\small hanwangzhang@ntu.edu.sg,\quad qianrusun@smu.edu.sg\\}

\begin{document}
\maketitle
\begin{abstract}

Even when using large multi-modal foundation models, few-shot learning is still challenging---if there is no proper inductive bias,  it is nearly impossible to keep the nuanced class attributes while removing the visually prominent attributes that spuriously correlate with class labels. To this end, we find an inductive bias that the time-steps of a Diffusion Model (DM) can isolate the nuanced class attributes, i.e., as the forward diffusion adds noise to an image at each time-step, nuanced attributes are usually lost at an earlier time-step than the spurious attributes that are visually prominent. Building on this, we propose \underline{Ti}me-step \underline{F}ew-shot (TiF) learner. We train class-specific low-rank adapters for a text-conditioned DM to make up for the lost attributes, such that images can be accurately reconstructed from their noisy ones given a prompt. Hence, at a small time-step, the adapter and prompt are essentially a parameterization of only the nuanced class attributes. For a test image, we can use the parameterization to only extract the nuanced class attributes for classification. TiF learner significantly outperforms OpenCLIP and its adapters on a variety of fine-grained and customized few-shot learning tasks. Codes are in \scalebox{0.88}{\url{https://github.com/yue-zhongqi/tif}}.

\end{abstract}    
\section{Introduction}
\label{sec:1}

Multi-modal foundation models, \eg, CLIP~\cite{radford2021learning},  have recently demonstrated remarkable zero-shot performance in general visual classification tasks~\cite{cherti2023reproducible, gpt4}. They define a task by a suitable text prompt for each class (\eg, ``a photo of an aircraft" for the class ``aircraft"), and use the model to classify an image by selecting the class prompt with the highest ``image-prompt'' similarity. However, the zero-shot paradigm encounters limitations 
when one cannot find a proper prompt to accurately describe the class of interest. 
This is particularly true for niche and fine-grained categories whose names may not encapsulate the subtle visual features recognizable by the model (\eg, ``707-320 aircraft''), or for customized categories where it is impractical to fully describe their specification through text alone (\eg, a specific person). For these situations, a few-shot training set is necessary to define the classification on demand.  

\begin{figure}
    \centering
    \includegraphics[width=\linewidth]{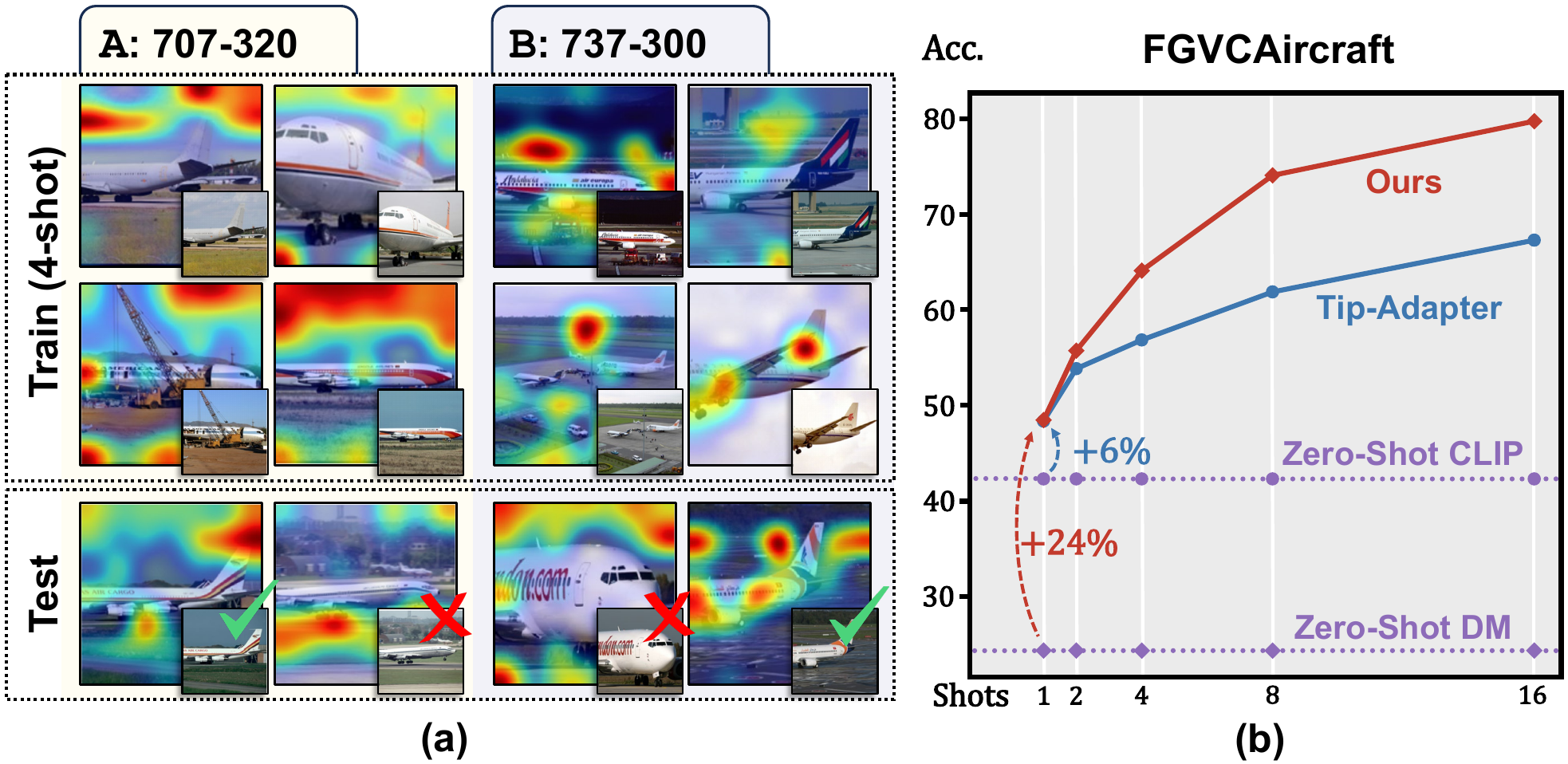}
    \caption{(a) GradCAM~\cite{selvaraju2017grad} of Tip-Adapter~\cite{tipadapter} on a 4-shot learning task from FGVCAircraft~\cite{fgvc}, where it is biased to the spurious background. (b) Comparison of few-shot learning performance. Our DM-based method significantly outperforms zero-shot OpenCLIP~\cite{ilharco_gabriel_2021_5143773} (ViT-H/14 trained on LAION-2B~\cite{schuhmann2022laionb}) and its adapter.}
    \label{fig:1}
    \vspace{-4mm}
\end{figure}

The most common approach is to train an auxiliary feature adapter attached after the CLIP visual encoder~\cite{clipadapter, tipadapter} or text encoder~\cite{coop, cocoop}. The objective is to align the adapted image feature with its prompt embedding only from the same class. However, it is well-known that such discriminative training on few-shot examples is biased to spurious correlations~\cite{yue2020interventional}, which hinders generalization at test time. For example, in Figure~\ref{fig:1}, consider a scenario where aircraft \texttt{A} and \texttt{B} is defined by a subtle class attribute, \eg, \emph{window}, but a visually prominent attribute happens to spuriously correlate with the class labels in the few-shot training set, \ie, \texttt{A} appearing with background \emph{sky} and \texttt{B} with \emph{ground}. The model will erroneously consider both \emph{window} and background attributes for inaccurate prediction, \eg, predicting \texttt{B} with \emph{sky} as \texttt{A}.
Frustratingly, it is impossible to isolate the nuanced class attributes from visually prominent yet spurious ones without an explicit inductive bias (\eg, prior knowledge that ``windows'' is a class attribute)~\cite{d2019multi, locatello2019challenging}.

To address the challenge, we turn our attention to generative classifiers~\cite{ng2001discriminative}, inspired by the recent advancement in the text-conditioned generative Diffusion Model (DM)~\cite{sd}. DM has enabled extrapolation to novel attribute combinations by textual control (\eg, generating ``Salvador Dalí with a robotic half-face''). Hence ideally, one could prompt DM with all attribute combinations for each class (\eg, ``\texttt{B} with sky'' and ``\texttt{B} with ground'') to synthesize a diverse training set, where no attribute is biased. Unfortunately, this approach is impractical yet, since there is no ground-truth of a complete attribute inventory.

In this paper, we introduce a practical solution by revealing that nuanced class attributes and visually prominent ones are naturally isolated by diffusion time-steps, lending itself to de-biasing.
Specifically, DM defines a forward process that gradually injects Gaussian noise into each image $\mathbf{x}_0$ over $T$ time-steps, creating a sequence of noisy images $\mathbf{x}_1,\ldots,\mathbf{x}_T$ that progressively collapse to pure noise. As $t$ increases, we show in Figure~\ref{fig:2} (top) that more visual attributes are \emph{lost} when they become indistinguishable.
Notably, we prove in Section~\ref{sec:3.3} that nuanced class attributes (\eg, windows defining class \texttt{A}, \texttt{B}) are lost at an early time-step, while visually prominent ones (\eg, the spurious background) are lost later.
This motivates our approach:

\begin{figure}
    \centering
    \includegraphics[width=\linewidth]{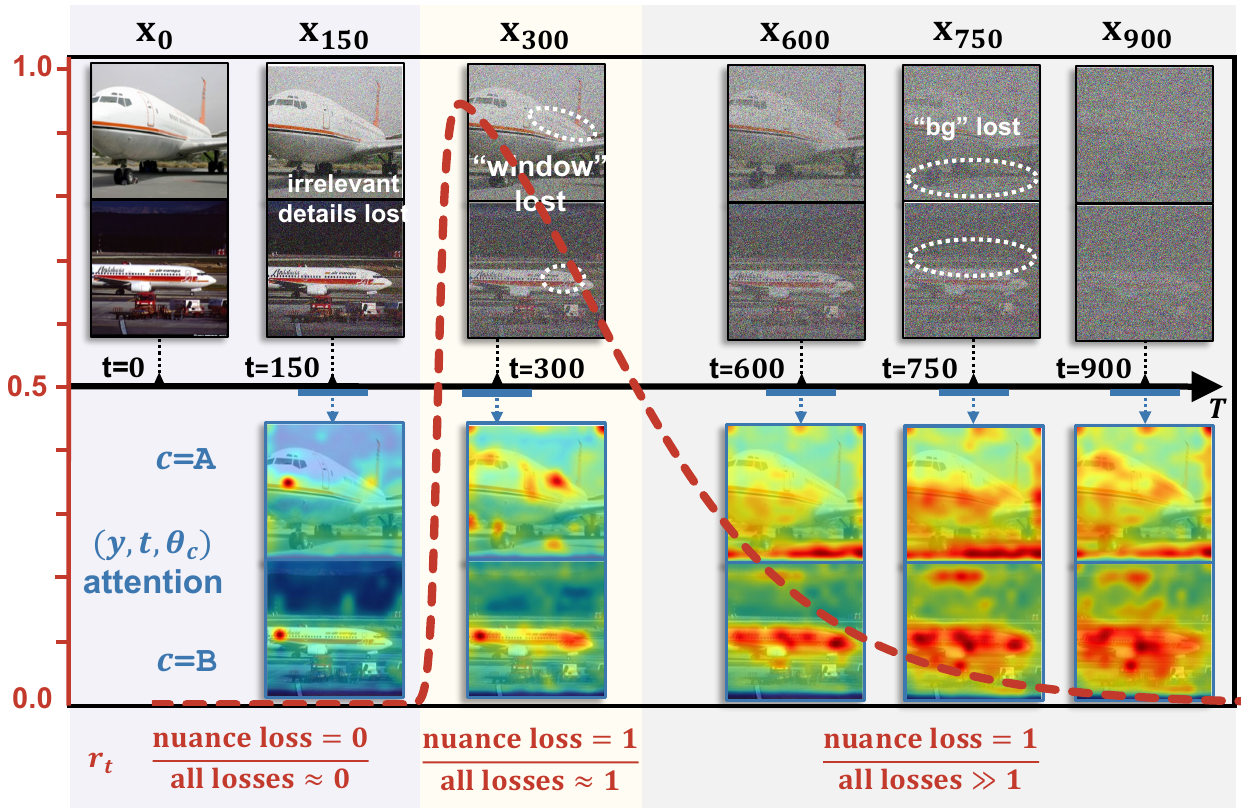}
    \caption{Top: DM forward process with attribute loss examples. Bottom: Attention map of what $(y,t,\theta_c)$ parameterizes for $c=$ \texttt{A} or \texttt{B}, which includes only nuances at a small $t$, and expands as $t$ increases when more attributes are lost. We follow~\cite{mokady2023null} to compute the average attention over a small time-step range indicated by the blue line. Details in Appendix. \textcolor{red}{Red}: Our proposed hyper-parameter-free weights for all time-steps.}
    \label{fig:2}
    \vspace{-4mm}
\end{figure}

\noindent\textbf{(\rom{1}) Parameterization}. For each class $c$, we train a denoising network $d$ parameterized by $\theta_c$ to reconstruct each image $\mathbf{x}_0$ in its training set from the noisy $\mathbf{x}_t$ and a text prompt $y$ (details given below), \ie, by minimizing the reconstruction error $\lVert d(\mathbf{x}_t, y, t; \theta_c) - \mathbf{x}_0 \rVert^2$ for all $t$.
When training converges with accurate reconstruction, $d(\cdot;\theta_c)$ must make up for the lost attributes in each $\mathbf{x}_t$ using $y$. Hence $(y, t, \theta_c)$ is essentially a \emph{parameterization} of the lost attributes at $t$ for class $c$.
Particularly, we adopt a low-rank implementation of $\theta_c$, such that $(y, t, \theta_c)$ at a small $t$ is constrained to parameterize only the lost nuances of $c$, without accidentally capturing the visually prominent attributes that are not lost yet.
We visualize this in Figure~\ref{fig:2} bottom by attention map, \eg, $(y, t, \theta_c)$ indeed parameterizes the visual attribute about window at $t=300$ when it becomes lost.

\noindent\textbf{(\rom{2}) De-biasing by Time-steps}. We classify a test image $\mathbf{x}_0$ by selecting $c$ with the least reconstruction error $\lVert d(\mathbf{x}_t, y, t; \theta_c) - \mathbf{x}_0 \rVert^2$ at a \emph{small time-step} $t$. This inference rule is not affected by the spurious correlations, because from the analysis above, such $(y, t, \theta_c)$ parameterize \emph{only} the nuanced attributes that define $\mathbf{x}_0$'s class, free from the visually prominent ones that are spurious.

We term our approach as the \underline{Ti}me-step \underline{F}ew-shot (\textbf{TiF}) learner. We implement the above two points as below:

\noindent\textbf{(\rom{1})} We inject Low-Rank Adaptation (LoRA) matrices~\cite{lora} into a pre-trained, frozen DM and denote their parameters as $\theta_c$. We follow DreamBooth~\cite{dreambooth} to use a rare token identifier [V] (\eg, ``hta'') to form the prompt $y$, such that it can be easily re-associated to the specificity of each class.

\noindent\textbf{(\rom{2})} To avoid searching for the best ``small $t$'', we design a hyper-parameter-free approach.
We first derive an equation that measures the degree of attribute loss in Section~\ref{sec:3.3}. Then we compute a ratio $r_t$ of nuanced attribute loss over all attribute losses at each $t$ to calculate a weighted reconstruction error $\sum_t r_t \lVert d(\mathbf{x}_t, y, t; \theta_c) - \mathbf{x}_0 \rVert^2$ for inference.
As shown in Figure~\ref{fig:2} red line, the weight reaches its peak when nuances become lost at a small $t$, and diminishes to 0 when more attributes become lost as $t$ increases.
Hence the weighted error accounts only for the inability of each $d(\cdot;\theta_c)$ in making up the nuanced class attribute to de-bias.

The contribution of the paper is summarized below.
\begin{itemize}[leftmargin=+0.2in,itemsep=3pt,topsep=3pt,parsep=0pt]
	\item We formulate few-shot learning (FSL) within the context of recent advancements in foundation models ( Section~\ref{sec:3.1}), and introduce a theoretical framework that isolates nuanced attributes from visually prominent ones by diffusion time-steps (Section~\ref{sec:3.3}).
	\item Motivated by our theoretical insights, we present a straightforward yet effective FSL approach that mitigates spurious correlations (Section~\ref{sec:4}).
	\item On fine-grained classification, Re-Identification and medical image classification, TiF learner significantly outperforms the powerful OpenCLIP and its adapters in various few-shot settings by up to 21.6\%.
\end{itemize}

\section{Related Works}
\label{sec:2}

\noindent\textbf{Conventional FSL} typically adopts a pre-training, meta-learning and fine-tuning paradigm~\cite{chen2019closer, zhang2020deepemd}. The first stage aims to capture rich prior knowledge as a feature backbone~\cite{dhillon2019baseline}. The second stage trains the model on ``sandbox'' FSL tasks to tailor it for the target task, \eg, learning a classifier weight generator~\cite{gidaris2019generating}, a distance kernel function in $k$-NN~\cite{vinyals2016matching}, a feature space to better seprate the classes~\cite{vinyals2016matching,zhang2020deepemd}, or even an initialization of the classifier~\cite{finn2017model}. The final stage involves training a classifier on the few-shot examples. However, multi-modal foundation models already capture extremely profound prior knowledge, hence recent works focus on few-shot adapting such models.

\noindent\textbf{FSL with Foundation Models}. There are two main approaches that both leverage CLIP. First is prompt tuning, which aims to learn a prompt for each class. CoOp~\cite{coop} learns a continuous prompt embedding instead of using a hand-crafted prompt. CoCoOp~\cite{cocoop} extends CoOp by learning an image conditional prompt. ProGrad~\cite{prograd} aligns the prompt gradient to the general knowledge of CLIP. Recent MaPLe~\cite{maple} additionally fine-tune the CLIP visual encoder. The other line aims to learn a CLIP visual feature adapter. CLIP-Adapter~\cite{clipadapter} applies a lightweight residual-style adapter, followed by the training-free approach Tip-Adapter~\cite{tipadapter}. CALIP~\cite{guo2022calip} proposes a parameter-free attention to improve both zero-shot and few-shot performance. The recent CaFo~\cite{zhang2023prompt} ensembles multiple foundation models to help with feature adaptation. However, they still suffer from the spurious correlation. Besides CLIP-based approaches, recent works have explored in-context learning with vision-language models~\cite{chen2023manipulating}, yet their current classification accuracy still lags behind.

\noindent\textbf{Alleviating Spurious Correlation}. Previous works use knowledge from additional data. For example, IFSL~\cite{yue2020interventional} leverages the data in pre-training, or unsupervised domain adaptation use unlabeled data in test domain~\cite{tang2020unsupervised,liu2021cycle,yue2023make}. We leverage the time-steps of DM without such data.

\section{Problem Formulations}
\label{sec:3}

\subsection{Few-Shot Learning}
\label{sec:3.1}

We aim to solve a $K$-way-$N$-shot Few-Shot Learning (FSL) task: train a model to classify $K$ categories using the few-shot dataset $\mathcal{D}$, where each category $c\in\{1,\ldots,K\}$ has a small number of $N$ images.
In particular, we use the notation in causal representation learning~\cite{higgins2018towards, scholkopf2021toward}: each image $\mathbf{x}$ from $c$ is generated by $\Phi(\mathbf{c},\mathbf{e})$, where $\Phi$ is the generator, $\mathbf{c}$ denotes the class attribute that define the category $c$ (\eg, ``many windows'' for class ``707-320''), and $\mathbf{e}$ denotes other environmental attribute (\eg, ``sky background'').
Note that $c$ is a category index, and $\mathbf{c}$ denotes its defining attribute.
Hence the crux of FSL is to pinpoint $\mathbf{c}$ for classification and discard the irrelevant $\mathbf{e}$.

\noindent\textbf{Necessity of FSL}.
The zero-shot accuracy of multi-modal foundation models matches that of a fully-supervised model on common visual categories~\cite{cherti2023reproducible,gpt4} (\eg, ImageNet~\cite{russakovsky2015imagenet}), rendering FSL unnecessary on those tasks.
Specifically, such model takes an image $\mathbf{x}$ and a text prompt $y_c$ that describes $c$ as input, and outputs the similarity between $\mathbf{x}$ and $c$.
For example, $\textrm{CLIP}(\mathbf{x},c):=\mathrm{cos}\left( V(\mathbf{x}), T(y_c) \right)$, where $\mathrm{cos}(\cdot,\cdot)$ is the cosine similarity, $V,T$ denotes the CLIP visual and text encoder, respectively.
Therefore we can predict an image $\mathbf{x}$ in a zero-shot setting as follows
\begin{equation}
    \hat{c}= \mathop{\mathrm{arg}\,\mathrm{max}}_{c\in\{1,\ldots,K\}} \textrm{CLIP}(\mathbf{x},c).
    \label{eq:1}
\end{equation}
Hence we focus on scenarios where the zero-shot paradigm is limited---\emph{when $\mathbf{c}$ is about nuanced details} on fine-grained or customized categories, it is impractical to find a prompt $y_c$ recognizable by the model that encapsulates the intricacies of $\mathbf{c}$, thereby requiring a few-shot set to define $c$.

\noindent\textbf{Challenge of FSL}. The most direct approach appends a trainable network parameterized by $\theta$ to the CLIP visual encoder $V$, denoted as $V(\cdot;\theta)$, and optimizes:
\begin{equation}
    \mathop{\mathrm{min}}_\theta \sum_{(\mathbf{x},c)\in \mathcal{D}} \frac{ \mathrm{cos}\left( V(\mathbf{x};\theta), T(y_c) \right) }{ \sum_{c'=1}^K \mathrm{cos}\left( V(\mathbf{x};\theta), T(y_{c'}) \right) },
    \label{eq:2}
\end{equation}
which trains $V(\mathbf{x};\theta)$ to predict its class prototype $T(y_c)$.
However, such discriminative training on few-shot examples is easily biased to the spurious correlation between $\mathbf{e}$ and $\mathbf{c}$.
Considering an extreme one-shot case, $\mathbf{x}=\Phi(\mathbf{c},\mathbf{e})$ and $\mathbf{x}'=\Phi(\mathbf{c}',\mathbf{e}')$ from another class differ by $\mathbf{e}\neq \mathbf{e}'$. $V(\mathbf{x};\theta)$ will inevitably mistake $\mathbf{e},\mathbf{e}'$ as part of the class attribute (\eg, predicting with both ``window'' and ``background'' in Figure~\ref{fig:1}), hence fail to reliably classify images not following the spurious pattern, \ie, $\Phi(\mathbf{c},\mathbf{e}')$ or $\Phi(\mathbf{c}',\mathbf{e})$.

\noindent\textbf{Time-step Prior and Limitation}. Our proposed TiF learner aims to circumvent the above challenge with the prior of Diffusion Model (DM)~\cite{ho2020denoising} time-steps: When the spurious $\mathbf{e}$ \emph{has a larger pixel-level impact} (\ie, visually prominent) than the nuanced class attribute $\mathbf{c}$, we can leverage the time-steps, introduced in Section~\ref{sec:3.2}, to isolate $\mathbf{c}$ at a small time-step in Section~\ref{sec:3.3}.
However there are two limiting cases:
\emph{1) when a fine-grained $\mathbf{e}$ spuriously correlates with $\mathbf{c}$}, it will also be isolated at a small time-step, hence requiring additional prior to remove it;
\emph{2) when a class attribute $\mathbf{c}$ is coarse-grained}, this becomes a hierarchical classification task out of this paper's scope. Note that this is also unlikely on our evaluation datasets (Section~\ref{sec:5.1}).

\subsection{Diffusion Model}
\label{sec:3.2}

DM is a generative model that first adds noise to images, and then learns to reconstruct them by a denoising network.

\noindent\textbf{Forward Process}. It adds Gaussian noise to each image $\mathbf{x}_0$ in $T$ time-steps, producing a sequence of noisy images $\mathbf{x}_1,\ldots,\mathbf{x}_T$, with the subscript denoting the time-step. Given $\mathbf{x}_0$ and a variance schedule $\beta_1,\ldots,\beta_T$ (\ie, how much noise is added at each time-step), $\mathbf{x}_t$ adheres to the following noisy sample distribution:
\begin{equation}
    q(\mathbf{x}_t | \mathbf{x}_0) = \mathcal{N} (\mathbf{x}_t; \sqrt{\bar{\alpha}_t} \mathbf{x}_0, (1-\bar{\alpha}_t) \mathbf{I}),
    \label{eq:3}
\end{equation}
where $\alpha_t:=1-\beta_t, \; \bar{\alpha}_t:=\prod_{s=1}^t \alpha_s$.

\noindent\textbf{Training}. DM learns a denoising network $d$ by first sampling a noisy image $\mathbf{x}_t$ at a random time-step $t$ from the distribution in Eq.~\eqref{eq:3}, and then training $d$ to
minimize the loss $\mathcal{L}_t$ of reconstructing $\mathbf{x}_0$ from $\mathbf{x}_t$:
\begin{equation}
    \mathcal{L}_t(d,\mathbf{x}_0,y) = w_t \mathop{\mathbb{E}}_{q(\mathbf{x}_t|\mathbf{x}_0)}  \lVert \mathbf{x}_0 - d (\mathbf{x}_t , y, t) \rVert^2,
    \label{eq:4}
\end{equation}
where $w_t$ is a standard weight (see Appendix), and $y$ is an optional condition, \eg, for an unconditional DM~\cite{ho2020denoising}, $y=\emptyset$; for a text-conditioned DM~\cite{sd}, $y$ is a text prompt describing $\mathbf{x}_0$.
Next, we show that nuanced $\mathbf{c}$ can be separated from visually prominent $\mathbf{e}$ by diffusion time-steps.

\subsection{Theory}
\label{sec:3.3}

\begin{figure}
    \centering
    \includegraphics[width=\linewidth]{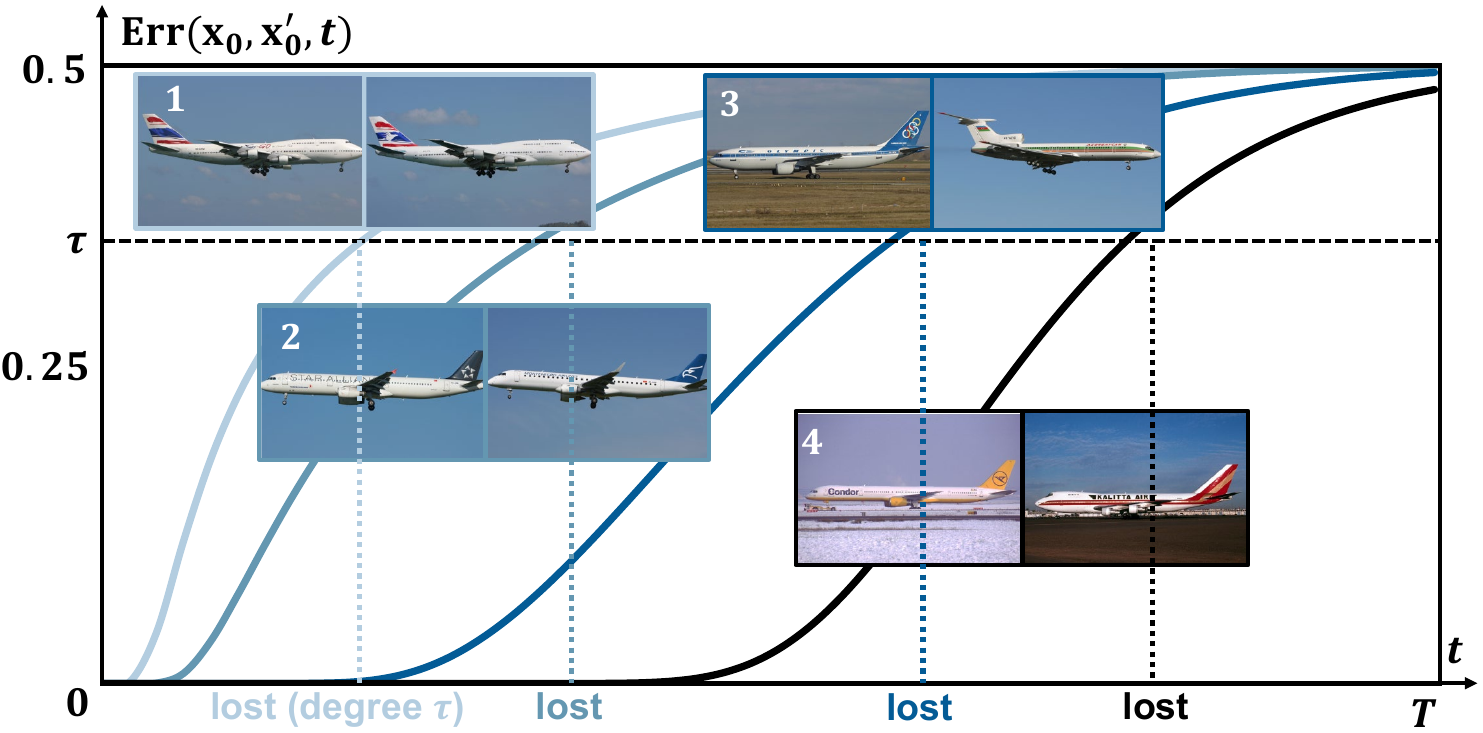}
    \caption{Plot of $\mathrm{Err}(\mathbf{x}_0,\mathbf{x'_0},t)$ on 4 pairs of $(\mathbf{x}_0,\mathbf{x}'_0)$ with different pixel-level differences. We observe that the attribute loss for each pair is strictly increasing in $t$, and the fine-grained attribute that distinguishing more similar image pair is lost earlier.}
    \label{fig:3}
    \vspace{-4mm}
\end{figure}

We prove that in the forward diffusion process, each fine-grained $\mathbf{c}\in \mathcal{C}$ is lost at an earlier time-step compared to more coarse-grained $\mathbf{e}\in \mathcal{E}$, where $\mathcal{C},\mathcal{E}$ denotes the set of all class attributes and environmental ones, respectively, and the granularity is defined by the pixel-level changes when altering an attribute.
We first formalize the notion of attribute loss, \ie, when an attribute becomes indistinguishable in the noisy images sampled from $q(\mathbf{x}_t|\mathbf{x}_0)$.

\noindent\textbf{Definition}. (Attribute Loss) \textit{Without loss of generality, we say that the attribute $\mathbf{c}\in\mathcal{C}$ is lost with degree $\tau$ at $t$ when $\mathop{\mathbb{E}}_{(\mathbf{c}', \mathbf{e}) \in \mathcal{C} \times \mathcal{E}} \left[ \mathrm{Err}\left( \Phi(\mathbf{c}, \mathbf{e}), \Phi(\mathbf{c}', \mathbf{e}), t\right) \right] \geq \tau$, where $\mathrm{Err}(\mathbf{x}_0, \mathbf{x}'_0, t)$ is defined as}
\begin{equation*}
    \mathop{\mathrm{min}}_d  \frac{1}{2} \left[ \mathop{\mathbb{E}}_{q(\mathbf{x}_t|\mathbf{x}_0)}  \mathbbm{1}\left( d(\mathbf{x}_t) = \mathbf{x}'_0 \right) 
    + \mathop{\mathbb{E}}_{q(\mathbf{x}'_t|\mathbf{x}'_0)} \mathbbm{1}\left( d(\mathbf{x}'_t) = \mathbf{x}_0 \right) \right],
\end{equation*}
\textit{with $d:\mathcal{X}\to\mathcal{X}$ and $\mathbbm{1}(\cdot)$ denoting the indicator function. We similarly define the degree of loss for each environmental attribute $\mathbf{e} \in \mathcal{E}$.}

Intuitively, $\mathrm{Err}(\mathbf{x}_0, \mathbf{x}'_0, t)$ measures the smallest error for a network $d$ to reconstruct the original image given noisy ones drawn from $q(\mathbf{x}_t|\mathbf{x}_0)$ or $q(\mathbf{x}'_t|\mathbf{x}'_0)$.
Hence, when $\mathbf{x}'_0$ differs from $\mathbf{x}_0$ only by $\mathbf{c}\neq\mathbf{c}'$, a larger $\mathrm{Err}(\mathbf{x}_0, \mathbf{x}'_0, t)$ means that the attribute $\mathbf{c}$ becomes harder to distinguish from $\mathbf{c}'$, \ie, more severe attribute loss.
Particularly, we prove the close-form of $\mathrm{Err}(\mathbf{x}_0, \mathbf{x}'_0, t)$ in Appendix: 
\begin{equation}
    \mathrm{Err}(\mathbf{x}_0, \mathbf{x}'_0, t) = \frac{1}{2} \left[ 1 - \mathrm{erf}\left(\frac{\lVert \sqrt{\bar{\alpha}_t} (\mathbf{x}_0-\mathbf{x}'_0) \rVert}{2\sqrt{2(1-\bar{\alpha}_t)}}\right) \right],
    \label{eq:5}
\end{equation}
where $\mathrm{erf}(z)=\frac{2}{\sqrt{\pi}} \int_{0}^z e^{-t^2}dt$ denotes the error function.
Hence we can compute attribute loss at each $t$ \wrt the pixel-level changes $\lVert \mathbf{x}_0-\mathbf{x}'_0 \rVert$ (plotted in Figure~\ref{fig:3}), allowing us to derive a de-biasing strategy in Section~\ref{sec:4.2}.

\noindent\textbf{Theorem}. \textit{1) For each $\mathbf{c}\in \mathcal{C}$, there exists a smallest time-step $t(\mathbf{c})$, such that $\mathbf{c}$ is lost with at least degree $\tau$ at each $t \in \{t(\mathbf{c}),\ldots, T\}$. This also holds for each $\mathbf{e}\in \mathcal{E}$. 2) $\exists T, \{\beta_i\}_{i=1}^T$ such that $t(\mathbf{e}) > t(\mathbf{c})$ whenever $\lVert \Phi(\mathbf{c}',\mathbf{e}) - \Phi(\mathbf{c}',\mathbf{e}') \rVert$ is first-order stochastic dominant over $\lVert \Phi(\mathbf{c},\mathbf{e}') - \Phi(\mathbf{c}',\mathbf{e}') \rVert$ with $\mathbf{c}' \sim \mathcal{C},\mathbf{e}' \sim \mathcal{E}$ uniformly.}

Intuitively, the first part of the theorem states that a lost attribute will not be regained as time-step $t$ increases, and there is a time-step $t(\mathbf{c})$ when $\mathbf{c}$ becomes lost (with degree $\tau$) for the first time.
The second part states that when changing the environmental attribute $\mathbf{e}$ is more likely to cause a larger pixel-level changes than changing the class attribute $\mathbf{c}$, then $\mathbf{e}$ becomes lost at a larger time-step compared to $\mathbf{c}$.
See Figure~\ref{fig:3} for an illustration of the two parts.
We build on this mechanism in Section~\ref{sec:4} to removes the visually prominent, yet spurious $\mathbf{e}$ for de-biasing in FSL.

\section{Approach}
\label{sec:4}

We have shown in Eq.~\eqref{eq:5} that attribute loss prevents a denoising network to accurately reconstruct $\mathbf{x}_0$ from the noisy $\mathbf{x}_t$.
Hence by contra-position, if we enable accurate reconstruction by minimizing Eq.~\eqref{eq:4} using a condition $y$, then the trained network $d$ must associate $y$ to the lost attributes to make up for their loss (Section~\ref{sec:4.1}).
Moreover, our theorem indicates that only the nuanced $\mathbf{c}$ is lost at a small time-step $t$, hence $y$ is only associated to $\mathbf{c}$ at $t$, allowing us to remove $\mathbf{e}$ to de-bias (Section~\ref{sec:4.2}).
We detail our TiF learner below.

\begin{figure*}
    \centering
    \includegraphics[width=\linewidth]{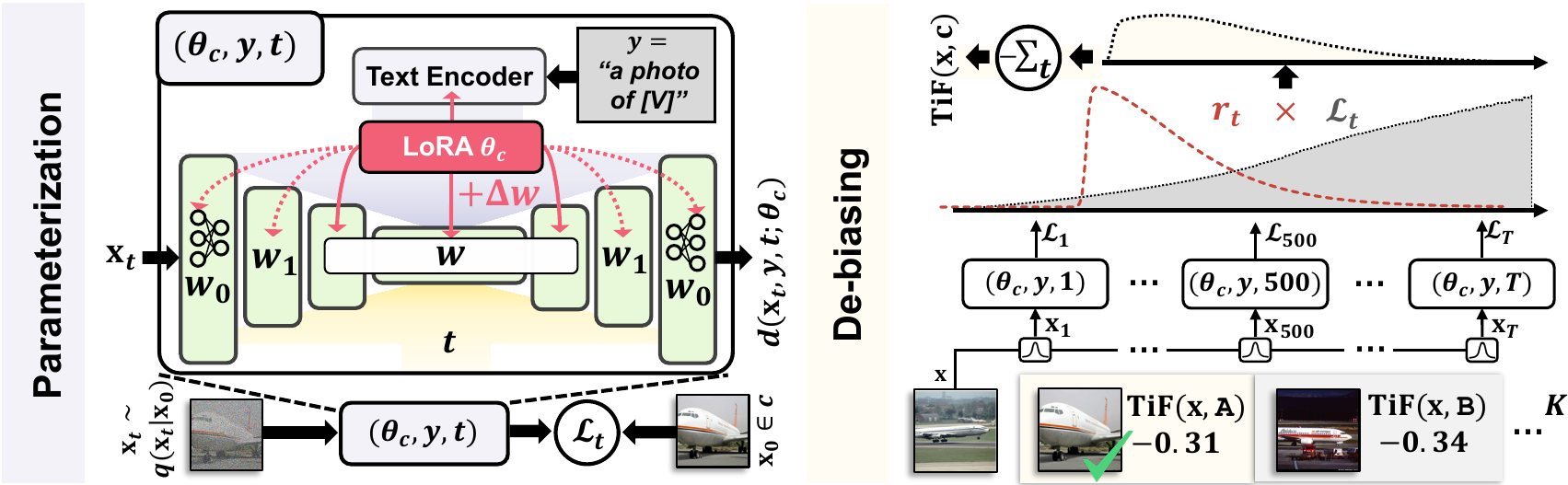}
    \caption{Overall pipeline of TiF learner. (a) Green: SD U-Net with each attention block illustrated by a rectangle. Red arrows: We inject trainable LoRA matrices $\theta_c$ to the attention blocks of U-Net and text encoder. Solid lines: always injected; dotted lines: optional (studied in ablation). We train $\theta_c$ to reconstruct $\mathbf{x}_0$ from $\mathbf{x}_t$ by $\mathcal{L}_t$. (b) Our inference rule by computing a weighted average $\mathcal{L}_t$ over time-steps.}
    \label{fig:method}
    \vspace{-4mm}
\end{figure*}

\subsection{Parameterization}
\label{sec:4.1}

We use the reconstruction loss in Eq.~\eqref{eq:4} to train a denoising network $d$ parameterized by $\theta_c$ on the few-shot examples of each category $c\in\{1,\ldots,K\}$ by:
\begin{equation}
    \mathop{\mathrm{min}}_{\theta_1,\ldots,\theta_K} \sum_{(\mathbf{x}_0,c) \in \mathcal{D}} \sum_{t=1}^T \mathcal{L}_t(d(\cdot; \theta_c), \mathbf{x}_0, y).
    \label{eq:6}
\end{equation}
When training converges, $(\theta_c,y,t)$ essentially parameterizes the lost attributes of category $c$ at each time-step $t$. We detail our implementation below:

\noindent\textbf{Choice of $d$}. We use the text-conditioned Stable Diffusion (SD)~\cite{sd}, as it is trained on internet-scaled data with extensive knowledge of visual attributes.
As shown in Figure~\ref{fig:method}, $d$ consists of a text encoder that extracts the text embedding from the prompt $y$, and a U-Net~\cite{ronneberger2015u} that leverages text and time-step embedding to reconstruct $\mathbf{x}_0$ from $\mathbf{x}_t$.
Note that SD encodes all images to a latent space. For convenience, we refer to the latent space when we say \textit{image}, $\mathbf{x}$ or \textit{pixel} with a slight abuse of notation.

\noindent\textbf{Choice of $\theta_c$}. As shown in Figure~\ref{fig:method}, instead of fine-tuning all the parameters of $d$, we freeze the pre-trained weights of $d$ and inject trainable Low-Rank Adaptation (LoRA) matrices to the linear layers in the text encoder and U-Net. For the U-Net, we use a subset of its attention blocks for injecting LoRA (ablation in Section~\ref{sec:5.3}). For each linear layer with weight $\mathbf{w}$, the new weight becomes $\mathbf{w}+\Delta\mathbf{w}$ after injection, where $\Delta\mathbf{w}$ is the injected low-rank matrix.
We denote the LoRA matrices parameters for category $c$ as $\theta_c$.
This offers several benefits:
1) The low-rank parameterization limits the model expressiveness to capture only the necessary attributes, \eg, only $\mathbf{c}$ at a small $t$ when $\mathbf{e}$ is yet lost.
2) As LoRA only affects the attention blocks, the model can only attend to the FSL task using existing visual knowledge, mitigating catastrophic forgetting~\cite{kirkpatrick2017overcoming} when adapting to the few-shot examples.
3) This also leads to efficient training and light-weight storage of $\theta_c$.

\noindent\textbf{Choice of $y$}. We use a fixed text prompt $y$=``a photo of [V]'' for all categories, where [V] is a rare token identifier following DreamBooth~\cite{dreambooth}. The intuition is to choose [V] with a weak semantic prior, such that $d(\cdot;\theta_c)$ does not need to detach it from its existing meaning first, before associating it with the specificity of category $c$. Note, it is not necessary to use a different $y$ for each $c$, as the LoRA matrices $\theta_c$ are class-specific and trained separately in Eq.~\eqref{eq:6}.

\subsection{De-biasing by Time-steps}
\label{sec:4.2}

\noindent\textbf{Inference Rule}. After training, the parameterization $(\theta_c,y,t)$ enables the denoising network $d$ to make up for the lost attributes at $t$ for category $c$. For a test image $\mathbf{x}$, our TiF learner defines its similarity with a category $c$ as the ability of $(\theta_c,y,t)$ to make up for the lost \emph{class attributes} (\ie, nuances) in $\mathbf{x}_t\sim q(\mathbf{x}_t | \mathbf{x})$ at each $t$, given by:
\begin{equation}
    \mathrm{TiF}(\mathbf{x},c)=-\sum_t r_t \mathcal{L}_t(d(\cdot;\theta_c),\mathbf{x},y),
\end{equation}
which enables inference by Eq.~\eqref{eq:1}. Here, $r_t$ is a ratio, namely, the degree of class attribute loss over that of all attribute losses. As shown in Figure~\ref{fig:2}, $r_t$ highlights the ability of $d(\cdot;\theta_c)$ to make up the loss of fine-grained $\mathbf{c}$ at a small $t$, and essentially disregards the ability to make up the visually prominent $\mathbf{e}$ by reaching towards 0 at a larger $t$. Hence TiF learner essentially mitigates the influence from $\mathbf{e}$ to de-bias. We show how to compute $r_t$ below.

\noindent\textbf{Degree of Class Attribute Loss}. To compute the degree of class attribute loss using Eq.~\eqref{eq:5}, we need to estimate the average pixel-level changes $\delta^*$ when \emph{only} altering the class attributes (measured by L2 distance).
A naive way is to directly find the minimum L2 distance between any two images from different categories. However, it is unlikely for the few-shot training set $\mathcal{D}$ to contain two images that differ only in $\mathbf{c}$, \eg, we may have class \texttt{A} on sky, \texttt{B} on ground and \texttt{C} with half sky and half ground. To this end, we estimate $\delta^*$ by a pixel-level approach:
\begin{equation}
    \delta^* = \sqrt{ \sum_{i=1}^W \sum_{j=1}^H \mathop{\mathrm{min}} \left\{ \lVert \mathbf{x}_{i,j} - \mathbf{x}'_{i,j} \rVert^2 \mid c \neq c' \right\} },
\end{equation}
where $(W,H)$ denotes the image size, and $\mathbf{x}_{i,j}$ is the pixel-level value at spatial location $(i,j)$. In the earlier example, the sky and ground of \texttt{C} can be matched by $\mathrm{min}$ with that of \texttt{A} and \texttt{B}, respectively, hence minimizing the influence of background.

\begin{table*}[t!]
\centering
\captionsetup{font=footnotesize,labelfont=footnotesize,skip=5pt}
\scalebox{0.8}{
\begin{tabular}{p{0.8cm}p{2.7cm} p{0.2cm}<{\centering} p{1.2cm}<{\centering}p{1.2cm}<{\centering}p{1.2cm}<{\centering}p{1.2cm}<{\centering}p{1.2cm}<{\centering} p{0.2cm}<{\centering} p{1.2cm}<{\centering}p{1.2cm}<{\centering}p{1.2cm}<{\centering}p{1.2cm}<{\centering}p{1.2cm}<{\centering}}
\hline\hline
\multirow{2}{*}{}  & \multicolumn{1}{c}{\multirow{2}{*}{\large{Method}}} &     & \multicolumn{5}{c}{\textbf{FGVCAircraft}} & &   \multicolumn{5}{c}{\textbf{ISIC2019}}\\ \cmidrule(lr){4-8}\cmidrule(lr){10-14}
& \multicolumn{1}{c}{}  &   & 1   & 2 & 4  &  8  & 16 &   &  1 & 2   & 4    & 8 & 16\\ \hline
\multicolumn{1}{c}{\multirow{4}{*}{\rotatebox{90}{\textbf{CLIP}}}} & Zero-Shot~\cite{radford2021learning} & & \multicolumn{5}{c}{24.9} & & \multicolumn{5}{c}{12.5} \\
    & CoOp~\cite{coop} & & 22.8\textcolor{black}{\scalebox{.7}{{-2.1}}} & 28.4\textcolor{black}{\scalebox{.7}{{+3.5}}} & 32.4\textcolor{black}{\scalebox{.7}{{+7.5}}} & 37.7\textcolor{black}{\scalebox{.7}{{+12.8}}} & 40.5\textcolor{black}{\scalebox{.7}{{+15.6}}} & & 10.8\textcolor{black}{\scalebox{.7}{{-1.7}}} & 19.3\textcolor{black}{\scalebox{.7}{{+6.8}}} & 19.6\textcolor{black}{\scalebox{.7}{{+7.1}}} & 24.2\textcolor{black}{\scalebox{.7}{{+11.7}}} & 25.8\textcolor{black}{\scalebox{.7}{{+13.3}}} \\
    & Co-CoOp~\cite{cocoop} & & 30.1\textcolor{black}{\scalebox{.7}{{+5.2}}} & 31.6\textcolor{black}{\scalebox{.7}{{+6.7}}} & 33.6\textcolor{black}{\scalebox{.7}{{+8.7}}} & 37.3\textcolor{black}{\scalebox{.7}{{+12.4}}} & 38.2\textcolor{black}{\scalebox{.7}{{+13.3}}} & & 11.4\textcolor{black}{\scalebox{.7}{{-1.1}}} & 13.9\textcolor{black}{\scalebox{.7}{{+1.4}}} & 16.6\textcolor{black}{\scalebox{.7}{{+4.1}}} & 21.8\textcolor{black}{\scalebox{.7}{{+9.3}}} & 22.8\textcolor{black}{\scalebox{.7}{{+10.3}}} \\
    & MaPLe*~\cite{maple} & & 30.1\textcolor{black}{\scalebox{.7}{{+5.2}}} & 33.0\textcolor{black}{\scalebox{.7}{{+8.1}}} & 33.8\textcolor{black}{\scalebox{.7}{{+8.9}}} & 39.4\textcolor{black}{\scalebox{.7}{{+14.5}}} & 40.7\textcolor{black}{\scalebox{.7}{{+15.8}}} & & 11.8\textcolor{black}{\scalebox{.7}{{-0.7}}} & 14.5\textcolor{black}{\scalebox{.7}{{+2.0}}} & 18.0\textcolor{black}{\scalebox{.7}{{+5.5}}} & 22.6\textcolor{black}{\scalebox{.7}{{+10.1}}} & 26.8\textcolor{black}{\scalebox{.7}{{+14.3}}} \\
\hline
\multicolumn{1}{c}{\multirow{5}{*}{\rotatebox{90}{\textbf{OpenCLIP}}}} & Zero-Shot~\cite{radford2021learning} & &   \multicolumn{5}{c}{42.3} & &  \multicolumn{5}{c}{16.9} \\
    & Linear-probe~\cite{radford2021learning} & & 18.4\textcolor{black}{\scalebox{.7}{{-23.9}}} & 32.5\textcolor{black}{\scalebox{.7}{{-9.8}}} & 44.1\textcolor{black}{\scalebox{.7}{{+1.8}}} & 55.0\textcolor{black}{\scalebox{.7}{{+12.7}}} & 59.8\textcolor{black}{\scalebox{.7}{{+17.5}}} & & 12.4\textcolor{black}{\scalebox{.7}{{-4.5}}} & 14.5\textcolor{black}{\scalebox{.7}{{-2.4}}} & 17.7\textcolor{black}{\scalebox{.7}{{+0.8}}} & 20.1\textcolor{black}{\scalebox{.7}{{+3.2}}} & 21.6\textcolor{black}{\scalebox{.7}{{+4.7}}} \\
    & Tip-Adapter~\cite{tipadapter} & & 47.7\textcolor{black}{\scalebox{.7}{{+5.4}}} & 51.6\textcolor{black}{\scalebox{.7}{{+9.3}}} & 54.7\textcolor{black}{\scalebox{.7}{{+12.4}}} & 58.4\textcolor{black}{\scalebox{.7}{{+16.0}}} & 62.2\textcolor{black}{\scalebox{.7}{{+19.9}}} & & 22.6\textcolor{black}{\scalebox{.7}{{+5.7}}} & 25.3\textcolor{black}{\scalebox{.7}{{+8.4}}} & 23.6\textcolor{black}{\scalebox{.7}{{+6.7}}} & 31.1\textcolor{black}{\scalebox{.7}{{+14.2}}} & 33.9\textcolor{black}{\scalebox{.7}{{+17.0}}} \\
    & Tip-Adapter-F~\cite{tipadapter} & & 48.4\textcolor{black}{\scalebox{.7}{{+6.1}}} & 53.9\textcolor{black}{\scalebox{.7}{{+11.6}}} & 56.9\textcolor{black}{\scalebox{.7}{{+14.7}}} & 62.0\textcolor{black}{\scalebox{.7}{{+19.7}}} & 67.4\textcolor{black}{\scalebox{.7}{{+25.1}}} & & 21.4\textcolor{black}{\scalebox{.7}{{+4.5}}} & 22.3\textcolor{black}{\scalebox{.7}{{+5.4}}} & 26.8\textcolor{black}{\scalebox{.7}{{+9.9}}} & 34.4\textcolor{black}{\scalebox{.7}{{+17.5}}} & 40.3\textcolor{black}{\scalebox{.7}{{+23.4}}} \\
    & CaFo*~\cite{zhang2023prompt} & & \textbf{51.7}\textcolor{black}{\scalebox{.7}{{+9.4}}} & {54.6}\textcolor{black}{\scalebox{.7}{{+12.3}}} & {58.5}\textcolor{black}{\scalebox{.7}{{+16.2}}} & {63.2}\textcolor{black}{\scalebox{.7}{{+20.9}}} & {66.7}\textcolor{black}{\scalebox{.7}{{+24.4}}} & & - & - & - & - & - \\
\hline

\multicolumn{1}{c}{\multirow{3}{*}{\rotatebox{90}{\textbf{DM}}}} & Zero-Shot~\cite{li2023diffusion} & &  \multicolumn{5}{c}{24.3} & &  \multicolumn{5}{c}{11.7} \\

    & \textbf{TiF learner w/o $c$} & &  \cellcolor{mygray}40.4\textcolor{black}{\scalebox{.7}{{+16.1}}} &  \cellcolor{mygray}53.8\textcolor{black}{\scalebox{.7}{{+29.5}}} &  \cellcolor{mygray}\textbf{65.0}\textcolor{red}{\scalebox{.7}{\textbf{+40.7}}} &  \cellcolor{mygray}72.1\textcolor{black}{\scalebox{.7}{{+47.8}}} &  \cellcolor{mygray}\textbf{80.4}\textcolor{red}{\scalebox{.7}{\textbf{+56.1}}} & \cellcolor{mygray} & \cellcolor{mygray}{23.9}\textcolor{black}{\scalebox{.7}{{+12.2}}} &  \cellcolor{mygray}{26.7}\textcolor{black}{\scalebox{.7}{{+15.0}}} &  \cellcolor{mygray}{33.3}\textcolor{black}{\scalebox{.7}{{+21.6}}} &  \cellcolor{mygray}{36.5}\textcolor{black}{\scalebox{.7}{{+24.8}}} &  \cellcolor{mygray}{43.8}\textcolor{black}{\scalebox{.7}{{+32.1}}} \\
    
    & \textbf{TiF learner} & & \cellcolor{mygray}{48.5}\textcolor{red}{\scalebox{.7}{\textbf{+24.2}}} &  \cellcolor{mygray}\textbf{55.8}\textcolor{red}{\scalebox{.7}{\textbf{+31.5}}} &  \cellcolor{mygray}{64.2}\textcolor{black}{\scalebox{.7}{{+39.9}}} &  \cellcolor{mygray}\textbf{74.2}\textcolor{red}{\scalebox{.7}{\textbf{+49.9}}} &  \cellcolor{mygray}{79.9}\textcolor{black}{\scalebox{.7}{{+55.6}}} & \cellcolor{mygray} &  \cellcolor{mygray}\textbf{24.1}\textcolor{red}{\scalebox{.7}{\textbf{+12.4}}} &  \cellcolor{mygray}\textbf{27.6}\textcolor{red}{\scalebox{.7}{\textbf{+15.9}}} &  \cellcolor{mygray}\textbf{33.8}\textcolor{red}{\scalebox{.7}{\textbf{+22.1}}} &  \cellcolor{mygray}\textbf{37.2}\textcolor{red}{\scalebox{.7}{\textbf{+25.5}}} &  \cellcolor{mygray}\textbf{44.7}\textcolor{red}{\scalebox{.7}{\textbf{+33.0}}} \\
    
\hline \hline
\end{tabular}}
\caption{$N$-shot accuracies on FGVCAircraft and ISIC2019. The small number on the right indicates the absolute gain over its corresponding zero-shot model. MaPLe additionally tunes the visual encoder besides prompt tuning. CaFo leverages an ensemble of foundation models (see main text). Tip-Adapter has two variants, w/o fine-tuning and w/ fine-tuning (denoted as Tip-Adapter-F). We include results of additional FSL tasks in Appendix.}
\label{tab:1}
\vspace{-0.3cm}
\end{table*}

\noindent\textbf{Computing $r_t$}. After obtaining $\delta^*$, we compute $r_t$ by
\begin{equation}
    r_t = \frac{ 1 - \mathrm{erf}\left(\gamma_t \delta^* \right) }{\int_{\delta^*}^\infty \left[1 - \mathrm{erf}(\gamma_t \delta) \right]  d\delta },
    \label{eq:9}
\end{equation}
where $\gamma_t=\sqrt{\bar{\alpha}_t} / 2\sqrt{2(1-\bar{\alpha}_t)}$ is the weight term from Eq.~\eqref{eq:5}. In the denominator, we essentially use the integral to simulate diverse environmental attributes that are more coarse-grained than the nuanced class attributes. We include the computation details in Appendix.

\section{Experiments}
\label{sec:5}

\subsection{Settings}
\label{sec:5.1}

\noindent\textbf{Datasets}.
Our experiments were conducted on four challenging and diverse datasets about fine-grained or customized classes: For \textbf{fine-grained} classification, we used: \textit{1) FGVCAircraft}~\cite{fgvc} consists of aircraft images from 100 classes with subtle visual differences. \textit{2) ISIC2019}~\cite{combalia2019bcn20000,tschandl2018ham10000} includes a diverse collection of dermatoscopic images with 8 types of skin lesions, serving as a testbed for our model's performance in medical image analysis. For \textbf{customized} classification, we used re-IDentification (reID) datasets, where each class is a specific identity (\eg, a person or car): \textit{3) DukeMTMC-reID}~\cite{duke} contains 702 person IDs captured by 8 surveillance cameras (for evaluation), and we sub-sampled 150 IDs with the most test images for few-shot evaluation. \textit{4) VeRi-776}~\cite{veri,liu2016deep} contains 200 vehicle IDs under 20 cameras, providing a diverse range of angles and environmental conditions.

\noindent\textbf{Evaluation Details}. We followed the protocol in~\cite{coop,clipadapter} to directly adapt models on the $K$-way-$N$-shot few-shot training set. This setting is more versatile with two main differences from the conventional FSL: 1) We did not perform pre-training or meta-learning on a many-shot training set relating with the FSL task~\cite{chen2019closer,Wertheimer_2021_CVPR,finn2017model}. 2) Our $K$ ranges from 100 to 200, much larger than the restricting $K=5$ typical in the conventional setting~\cite{vinyals2016matching,snell2017prototypical}.
On FGVCAircraft, we sampled the few-shot set from its train split and evaluated accuracy on its test split. On ISIC2019, as its test split has no ground-truth labels, we sampled few-shot set from the train split and tested on the rest images. We used the macro F1 score to account for the class imbalance. On the reID datasets, we sampled few-shot images from their gallery set, and evaluated (rank-1) accuracy on their query set.

\noindent\textbf{Implementation Details}. We used SD 2.0 following~\cite{li2023diffusion}, as we empirically found that the more commonly used SD 2.1 performs slightly worse in classification (see Appendix). For LoRA matrices rank, we used 8 on ISIC2019 and 16 on the rest datasets by visual inspection of the generation quality (see Section~\ref{sec:5.3}). We used a fixed rare token identifier [V]=``hta'' for all experiments. 
On dataset with class names, zero-shot CLIP and its adapter uses ``a photo of [$c$], a type of [SC]'' for each prompt $y_c$, where [$c$] denotes the name for class $c$ and [SC] is a dataset-specific super-class name, \ie, ``aircraft'' on FGVCAircraft and ``skin lesion'' on ISIC2019. For fair comparison, we implemented our $y$ as ``a photo of [V] [$c$], a type of [SC]''. We also tried a variant without using class names (denoted as w/o $c$) by setting $y$ as ``a photo of [V], a type of [SC]''.
%
For inference, we modified the implementation in~\cite{li2023diffusion} to add our weight in Eq.~\eqref{eq:9}. Other details in Appendix.

\noindent\textbf{Baselines}. We used two CLIP variants: CLIP with ViT-B/16 backbone trained on 400M image-caption pairs, and OpenCLIP with ViT-H/14 trained on LAION-2B. We compared with two lines of methods: 1) CLIP-Adapter~\cite{clipadapter}, Tip-Adapter~\cite{tipadapter} and CaFo~\cite{zhang2023prompt} based on Eq.~\eqref{eq:2}.
2) Prompt-tuning methods CoOp~\cite{coop}, Co-CoOp~\cite{cocoop} and MaPLe~\cite{maple}, which aim to learn $y_c$ in Eq.~\ref{eq:2}. However, We did not test them on OpenCLIP, as their official implementations are specifically designed for CLIP and difficult to be fairly reproduced on OpenCLIP. We also compared with the zero-shot Diffusion Classifier~\cite{li2023diffusion}, which is based on SD.

\subsection{Main Results}
\label{sec:5.2}

\begin{table*}[t!]
\centering
\scalebox{0.8}{
\begin{tabular}{p{0.8cm}p{2.7cm} p{0.2cm}<{\centering} p{1.2cm}<{\centering}p{1.2cm}<{\centering}p{1.2cm}<{\centering}p{1.2cm}<{\centering}p{1.2cm}<{\centering} p{0.2cm}<{\centering} p{1.2cm}<{\centering}p{1.2cm}<{\centering}p{1.2cm}<{\centering}p{1.2cm}<{\centering}p{1.2cm}<{\centering}}
\hline\hline
\multirow{2}{*}{}  & \multicolumn{1}{c}{\multirow{2}{*}{\large{Method}}} &     & \multicolumn{5}{c}{\textbf{DukeMTMC-reID}} & &   \multicolumn{5}{c}{\textbf{VeRi-776}}\\ \cmidrule(lr){4-8}\cmidrule(lr){10-14}
& \multicolumn{1}{c}{}  &   & 1   & 2 & 4  &  8  & 16 &   &  1 & 2   & 4    & 8 & 16\\ \hline
\multicolumn{1}{c}{\multirow{3}{*}{\rotatebox{90}{\textbf{CLIP}}}} & CoOp~\cite{coop} & & 8.2 & 10.4 & 17.2 & 29.6 & 33.7 & & 11.4 & 13.9 & 18.1 & 30.3 & 34.1 \\
    & Co-CoOp~\cite{cocoop} & & 9.7 & 12.1 & 20.1 & 32.7 & 44.5 & & 30.1 & 31.6 & 33.6 & 37.3 & 38.2 \\
    & MaPLe~\cite{maple} & & 13.5 & 32.9 & 40.7 & 55.7 & 63.0 & & 35.2 & 40.7 & 44.4 & 57.5 & 68.1 \\
\hline
\multicolumn{1}{c}{\multirow{3}{*}{\rotatebox{90}{\textbf{OC}}}} & Linear-probe~\cite{radford2021learning} & & 11.9 & 13.2 & 37.7 & 53.5 & 60.2 & & 16.7 & 29.5 & 51.1 & 61.1 & 69.8 \\
    & Tip-Adapter~\cite{tipadapter} & & 28.5 & 36.9 & 46.5 & 59.3 & 66.7 & & 36.4 & 47.5 & 59.8 & 71.2 & 80.1 \\
    & Tip-Adapter-F~\cite{tipadapter} & & 29.3 & 32.0 & 54.4 & 74.2 & 82.6 & & 37.4 & 48.6 & 62.2 & 79.1 & 85.4 \\
\hline
\multicolumn{1}{c}{\multirow{1}{*}{\rotatebox{90}{}}} & \textbf{TiF learner w/o $c$} & &  \cellcolor{mygray}\textbf{36.9} &  \cellcolor{mygray}\textbf{53.6} &  \cellcolor{mygray}\textbf{73.1} &  \cellcolor{mygray}\textbf{83.7} &  \cellcolor{mygray}\textbf{91.6} & \cellcolor{mygray} &  \cellcolor{mygray}\textbf{41.9} &  \cellcolor{mygray}\textbf{60.7} &  \cellcolor{mygray}\textbf{78.2} &  \cellcolor{mygray}\textbf{91.2} &  \cellcolor{mygray}\textbf{96.8} \\
\hline \hline
\end{tabular}}
\caption{$N$-shot accuracies on DukeMTMC-reID and VeRi-776. OC is the short for ``OpenCLIP''. Note that the zero-shot methods are no longer applicable here, as there are no class names or ground-truth text description of the classes.}
\label{tab:2}
\vspace{-0.3cm}
\end{table*}
\begin{table*}[t!]
\centering
\scalebox{0.8}{
\begin{tabular}{p{0.8cm}p{2.7cm} p{0.2cm}<{\centering} p{1.2cm}<{\centering}p{1.2cm}<{\centering}p{1.2cm}<{\centering}p{1.2cm}<{\centering}p{1.2cm}<{\centering} p{0.2cm}<{\centering} p{1.2cm}<{\centering}p{1.2cm}<{\centering}p{1.2cm}<{\centering}p{1.2cm}<{\centering}p{1.2cm}<{\centering}}
\hline\hline
\multirow{2}{*}{}  & \multicolumn{1}{c}{\multirow{2}{*}{\large{Method}}} &     & \multicolumn{5}{c}{\textbf{FGVCAircraft}} & &   \multicolumn{5}{c}{\textbf{DukeMTMC-reID}}\\ \cmidrule(lr){4-8}\cmidrule(lr){10-14}
& \multicolumn{1}{c}{}  &   & 1   & 2 & 4  &  8  & 16 &   &  1 & 2   & 4    & 8 & 16\\ \hline
\multicolumn{1}{c}{\multirow{3}{*}{\rotatebox{90}{\textbf{Subset}}}} & last & &  \textbf{48.5} & \textbf{55.8} & 64.2 & 73.1 & 78.1  & &  \textbf{36.9} & \textbf{53.6} & \textbf{73.1} & 84.2 & 91.3  \\
    & last + $w_1$ & & 48.2 & 55.7 & {65.0} & \textbf{74.2} & 78.5 & & 28.7 & 47.7 & 69.7 & \textbf{84.5} & 91.1 \\
    & last + $w_1$+$w_0$ & & 46.5 & 55.4 & \textbf{65.1} & 74.0 & \textbf{79.9} & & 25.8 & 43.2 & 65.7 & 83.4 & \textbf{91.6} \\
\hline
\multicolumn{1}{c}{\multirow{3}{*}{\rotatebox{90}{\textbf{Weight}}}} & ELB weight & &   44.7 & 50.2 & 59.9 & 68.7 & 71.2  & & 33.2 & 50.2 & 68.5 & 78.3 & 81.8  \\
    & PDAE & & \textbf{49.2} & \textbf{56.7} & 64.1 & 72.2 & 78.1 & & 36.2 & \textbf{53.9} & \textbf{73.4} & 81.2 & 90.8 \\
    & Ours & & 48.5 & 55.8 & \textbf{64.2} & \textbf{74.2} & \textbf{79.9} & & \textbf{36.9} & 53.6 & 73.1 & \textbf{84.5} & \textbf{91.6} \\
\hline \hline
\end{tabular}}
\caption{Top: Ablation on the LoRA injection subset. Last stands for the last attention block of the SD U-Net (solid lines in Figure~\ref{fig:method}). ``+ $w_1$'' and ``+ $w_2$'' stands for injecting the corresponding attention blocks in Figure~\ref{fig:method}. Bottom: Ablation on the time-step weights in inference. See main text for details.}
\label{tab:3}
\vspace{-0.3cm}
\end{table*}

\noindent\textbf{Overall Results}. As shown in Table~\ref{tab:1} and~\ref{tab:2}, our TiF learner achieves state-of-the-art performance across fine-grained and reID datasets, \eg, significantly improving existing methods by 13.7\% on FGVCAircraft, 21.6\% on DukeMTMC-reID and 16\% on VeRi-776.
On ISIC2019, we still achieve up to 7\% absolute gain over the best-performing baseline, despite the challenges posed by diverse appearances within each class, as well as the stringent macro-F1 evaluation, which is sensitive to under-performing classes.
Note that prompt tuning methods generally have non-ideal performances. This holds especially on customized reID classes with no class name to provide semantic prior. This validates the difficulties to describe the specification of fine-grained classes by prompt alone.

\noindent\textbf{Low vs. High Shots}. On low-shot settings (\eg, 1- or 2-shot), we observe that our method may not improve significantly. However, not all foundation models are equal, and we must account for the discriminative capability of a zero-shot model. By checking the gain of each method over its zero-shot foundation model (small numbers) in Table~\ref{tab:1}, we observe that our method improves the most, \eg, by +24.2\% with just 1-shot on FGVCAircraft. We also highlight that CaFo leverages an ensemble of multiple models (\ie, CLIP, DALL-E, GPT-3 and DINO), which provides rich prior to enable higher low-shot performance. By increasing \#shots, our TiF learner's accuracy continue to grow strongly, \eg, reaching near perfect predicting on the challenging 200-way-16-shot VeRi-776 task. This validates that our method can reap the benefits from well-defined class attributes using more shots. Moreover, the overall large improvements over baselines on 16-shot also highlights the persistence of spurious correlations, \ie, any attribute from the diverse visual attributes set can spuriously correlate with $\mathbf{c}$ to challenge a few-shot learner.


\noindent\textbf{With vs. Without [$\mathbf{c}$]}. On fine-grained datasets, we have access to the class name, and can include the class name in our prompt (see implementation details). Hence we compared the performance of TiF learner when using or without using class names (w/o $\mathbf{c}$) in Table~\ref{tab:1}. On FGVCAircraft, using class names significantly improves the 1- and 2-shot settings. This is because the class names can provide semantic prior to help describe the classes, which are otherwise ill-defined given the extreme low number of training images. However, on ISIC2019, using class names do not help with low-shot settings as much, and we conjecture that the prior knowledge of SD on ISIC2019 is significantly weaker compared to that on FGVCAircraft (much lower zero-shot accuracy). On higher-shot settings, adding class names overall has little to no effect, as the expanded training set can properly define each class.

\subsection{Ablations}
\label{sec:5.3}

\noindent\textbf{LoRA Injection Subset}. In conventional few-shot learning works~\cite{sun2019meta}, it is common to train only the last few layers of the backbone, corresponding to low-level features. As our goal is to capture fine-grained attributes, we are motivated to inject LoRA to the last attention blocks of the SD U-Net, as shown in Figure~\ref{fig:method}. Starting from just injecting the last block (solid line), we tried expanding the injected blocks and show the results in Table~\ref{tab:3} top. On 1-, 2- and 4-shot settings, injecting last attention block brings consistent improvements over other options. We conjecture that this provides additional inductive bias for the model to focus on the fine-grained details (controlled by the last block). However, on 8-shot and 16-shot settings, this becomes suboptimal, and we postulate that this can limit the expressiveness of the injected SD, such that it fails to capture the specification of each class. Overall, ``last + $w_1$'' and ``last + $w_1$ + $w_2$'' work the best for 8- and 16-shot, respectively. Hence we use these settings for all experiments in Table~\ref{tab:1} and~\ref{tab:2}. See additional ablation on this in Appendix.

\noindent\textbf{Inference Weight $r_t$}. In Table~\ref{tab:3} bottom, we compared our inference weight $r_t$ in Eq.~\eqref{eq:9} with two other weights. The ELB weight is derived from the evidence lower bound of the training data distribution (see Appendix), which respects any spurious correlation, \eg, if most airplanes have sky background, the weight will encourage DM to follow this pattern in its generation. Hence it leads to sub-optimal performance when we want to suppress the spurious correlations in FSL. The PDAE weight~\cite{zhang2022unsupervised} is developed to help distill visual attribute knowledge from a pre-trained DM, and also penalizes large time-steps when distillation is empirically difficult. Hence overall, it also performs reasonably well. However, it comes with a hyper-parameter $\gamma$, and when we use its default setting $\gamma=0.1$, it does not perform well on some settings, \eg, 16-shot on FGVCAircraft. In contrast, our proposed hyper-parameter-free adaptive weight is overall more stable due to its tailored design to isolate nuanced attributes from visually prominent ones.

\begin{figure}
    \centering
    \includegraphics[width=\linewidth]{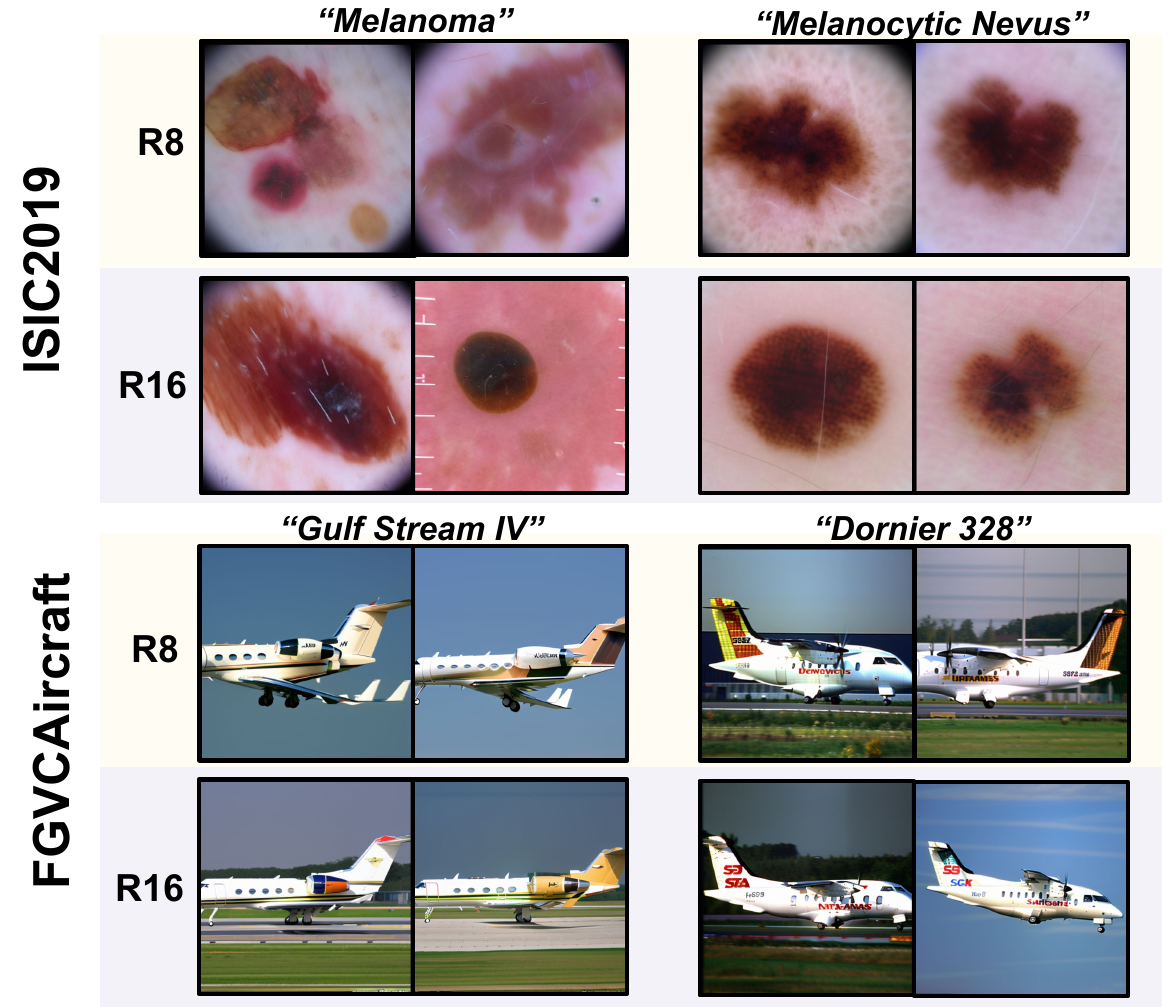}
    \caption{Comparison of synthesized images with two LoRA ranks. LoRA overfits to irrelevant details when rank is too high (top, rank 16), or fails to capture the nuances accurately when rank is too low (bottom, rank 8).}
    \label{fig:5}
    \vspace{-6mm}
\end{figure}

\noindent\textbf{LoRA Rank} is the rank of each LoRA matrix $\Delta \mathbf{w}$. As our method is based on a generative model, we can easily choose its rank through a visual approach on the train set. Specifically, we synthesize images using the LoRA-injected SD. A desired rank should enable SD to generate the specification of each class, but still limits it from overfitting to irrelevant details. As shown in Figure~\ref{fig:5}, rank 16 on ISIC2019 is too high, as the model overfits to the scratches on the lens, while rank 8 on FGVCAircraft is too low, as the model fails to capture the wing and the vertical stabilizer. Overall, we used rank 8 for ISIC2019 with simpler visual attributes and rank 16 for the rest.

\begin{figure}
    \centering
    \includegraphics[width=\linewidth]{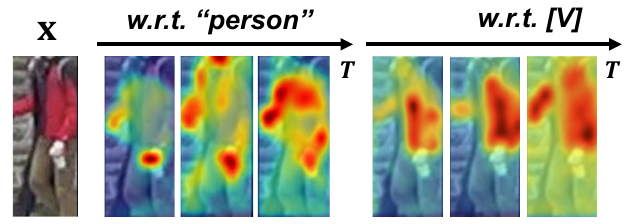}
    \caption{Attention map \wrt token ``person'' and ``[V]''.}
    \label{fig:7}
    \vspace{-6mm}
\end{figure}


\noindent\textbf{Additional Attention Maps}. In Figure~\ref{fig:7}, we show attention maps on DukeMTMC-reID \wrt different input tokens, where ``person'' corresponds to human parts and ``[V]'' is associated with the class attribute clothes that uniquely identify this person.

\section{Conclusions}

We presented TiF learner, a novel few-shot learner parameterization based on the Diffusion Model (DM), leveraging the inductive bias of its time-steps to isolate nuanced class attributes from visually prominent, yet spurious ones.
Specifically, we theoretically show that in the forward diffusion process, nuanced attributes are lost at a smaller time-step than the visually prominent ones.
Based on this, we train class-specific low-rank adapters that enables a text-conditioned DM to make up for the attribute loss by accurately reconstructing images from their noisy ones given a prompt. Hence each adapter and the prompt parameterizes only the nuanced class attributes at a small time-step, enabling a robust inference rule that focuses on the class attributes.
Extensive results show that our method significantly outperforms strong baselines on various fine-grained or customized few-shot learning tasks. As future direction, we will seek additional inductive bias to tackle more complicated scenarios (\eg, hierarchical classification).

\section{Acknowledgement}

This research is supported by the National Research Foundation, Singapore under its AI Singapore Programme (AISG Award No: AISG2-RP-2021-022), MOE AcRF Tier 2 (MOE2019-T2-2-062), Wallenberg-NTU Presidential Postdoctoral Fellowship, the A*STAR under its AME YIRG Grant (Project No.A20E6c0101) and the Lee Kong Chian (LKC) Fellowship fund awarded by Singapore Management University. Pan Zhou was supported by the Singapore Ministry of Education (MOE) Academic Research Fund (AcRF) Tier 1 grant.
{
    \small
    \bibliographystyle{ieeenat_fullname}
    \bibliography{main}
}


\end{document}